\pdfoutput=1
\documentclass[10pt]{article}
\usepackage{authblk}
\usepackage[final]{emnlp2022}
\usepackage{times}
\usepackage{latexsym}
\usepackage[T1]{fontenc}
\usepackage[utf8]{inputenc}
\usepackage{microtype}
\usepackage{inconsolata}
\usepackage{fullpage}
\usepackage{enumitem}
\usepackage{pgfplots}
\usepackage{algorithm}
\usepackage{algpseudocode}
\usepackage{graphicx}
\usepackage{placeins}
\usepackage{tabularx}
\usepackage{makecell}
\usepackage{booktabs}       
\usepackage{array}  
\usepackage{dblfloatfix}
\usepackage{verbatim}
\usepackage{colortbl}
\newcolumntype{P}[1]{>{\centering\arraybackslash}p{#1}}
\newcolumntype{M}[1]{>{\centering\arraybackslash}m{#1}}

\usepackage{subcaption}
\usepackage{amssymb}
\usepackage{amsmath}
\usepackage{mathtools}
\usepackage[normalem]{ulem}

\def\1{\bm{1}}

\definecolor{bblue}{HTML}{4F81BD}
\definecolor{rred}{HTML}{C0504D}
\definecolor{ggreen}{HTML}{9BBB59}
\definecolor{ppurple}{HTML}{9F4C7C}
\definecolor{oorange}{HTML}{FFA500}

\pgfplotsset{width=8cm,compat=1.9}
\graphicspath{ {./media/} }

\title{Sparse*BERT: Sparse Models Generalize To New tasks and Domains }

\author[1,2]{Daniel Campos\thanks{~~~Corresponding author: dcampos3@illinois.edu}}
\author[2]{Alexandre Marques}
\author[2]{Tuan Nguyen}
\author[2]{Mark Kurtz}
\author[1]{ChengXiang Zhai}
\affil[1]{Department of Computer Science, the University of Illinois Urbana-Champaign}
\affil[2]{Neural Magic Inc.}
\begin{document}
\maketitle
\begin{abstract}
Large Language Models have become the core architecture upon which most modern natural language processing (NLP) systems build. These models can consistently deliver impressive accuracy and robustness across tasks and domains, but their high computational overhead can make inference difficult and expensive. To make using these models less costly recent work has explored leveraging \textit{unstructured pruning} to improve inference speed and decrease size. This paper studies how models pruned using Gradual Unstructured Magnitude Pruning can transfer between domains and tasks. Our experimentation shows that pruned models using general domain masked language models during pretraining can transfer to new domains and tasks without extensive hyperparameter explorations. We demonstrate that our general sparse model \textit{Sparse*BERT} can specialize simply by pretraining the compressed architecture on unstructured biomedical text to become SparseBioBERT. SparseBioBERT can match and exceed the performance of BioBERT with 10\% of the parameters.
\end{abstract}
\section{Introduction}
Foundational Models \cite{Bommasani2021OnTO} based on the Transformer architecture \cite{Vaswani2017AttentionIA} has quickly become the most common building block in the modern language understanding stack, 
providing robust language representations which can be leveraged to provide impressive accuracy on tasks like question answering, text classification, and token classification. These Large Language Models (LLMs) are able to adapt to novel domains through pretraining resulting in models like BioBERT \cite{Lee2020BioBERTAP}, LegalBERT \cite{Chalkidis2020LEGALBERTTM}, and SciBERT \cite{beltagy2019SciBERTAP} have become a popular strategy for improving performance further. While accurate and robust, LLMs are not without drawbacks. 
They commonly have hundreds of millions or billions of parameters requiring large specialized computer clusters to run inference at scale. Several approaches have been successfully used to improve the performance of these LLMs, such as approximating attention \cite{Peng2021RandomFA}, removing portions of the models \cite{Sridhar2020UndividedAA}, and reducing the precision of activation and weight values.\\
\begin{figure}[!
t]
    \centering
    \includegraphics[scale=0.42
    ]{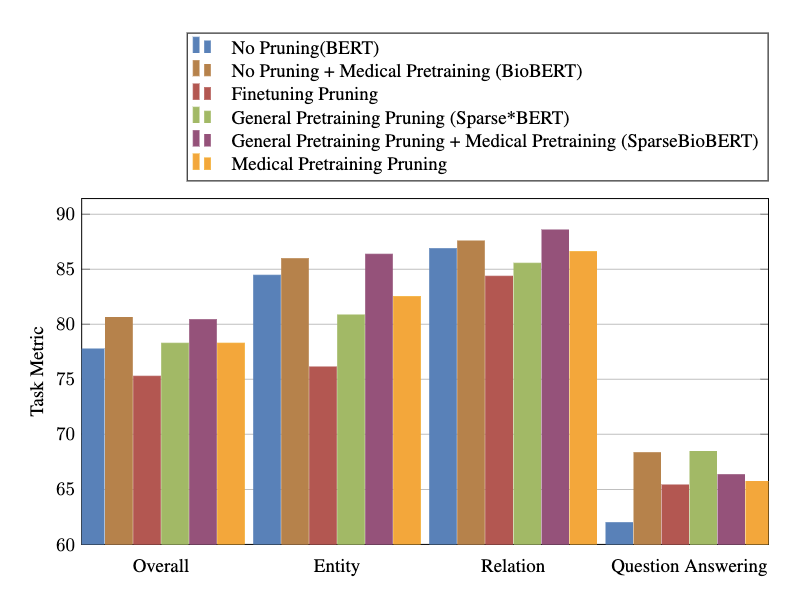}
    \vspace{-1.2em}
\caption{Impact of stage and domain of pruning when evaluating BERT base uncased on medical NLP Entity Extraction, Relation Extraction, and Question Answering. For pruned and unpruned models medical pretraining can improve performance, and the performance of the pruned model and the performance of BioBERT and SparseBioBERT is comparable.}
\label{fig:sparse_transfer_categorical}
\end{figure}
Recent work \cite{Zafrir2021PruneOF} \cite{Kurti2022TheOB} has shown that the application of unstructured and semi-structured (block) pruning mechanisms on LLMs can significantly compress models with little to no loss in accuracy. While these approaches are successful and applicable during model general domain pretraining and task-specific fine-tuning, prior work has not studied how pruned models transfer to new domains nor the impact of pretraining stage pruning on transfer accuracy. Given that most applications would require the transfer of the general domain LLMs to a specific application domain, it is important to study the generality and robustness of the pruned LLMs when applied to multiple tasks in an application domain.  \\
While existing pruning research has found it possible to prune models heavily without loss in accuracy, most approaches have focused on the compression of individual tasks or textual domains. These specialized models match or exceed the accuracy of the dense model but commonly require vast amounts of hyperparameter turning and task-specific optimization to achieve this result. Compressed models like DistillBERT \cite{Sanh2019DistilBERTAD} and TinyBERT \cite{Jiao2020TinyBERTDB} are some of the most popular LLMs because they provide compression without any additional know-how or optimization. For pruned models to become a common architecture for NLP tasks, they must be as robust and easy to use as their uncompressed counterparts. \\
This paper explores this potential and proposes Sparse*BERT, a new pruned LLM that can adapt effectively to new domains without extensive fine-tuning or task-specific pruning.\\
Our work studies generalizable pruned LLMs by evaluating how well they can transfer to previously unstudied tasks in the biomedical domain.
Specifically, we study these questions by focusing on transferring pruned and unpruned LLMs to the biomedical domain and evaluating the accuracy of said models on downstream tasks like Entity Extraction(EE), Relation Extraction(RE), and Question Answering(QA). Our experiments demonstrate that pruned LLMs generalize well and are robust to domain transfer and variation in target task, dataset size, and dataset difficulty. In summary, our contributions are as follows:
\begin{itemize} 
\item We reinforce Zafrir et al.'s findings that LLMs pruned on the general domain data can transfer to new domains without extensive hyperparameter tuning and extend their work, demonstrating these pruned models can be transferred to new pre-training domains without additional parameter optimization.
\item We introduce a sparse model, Sparse*BERT, and its domain adaptation adapted for the Medical/Bio NLP domain called SparseBioBERT. This model matches the accuracy of the BioBERT with $10\%$ of the active parameters.
\end{itemize}
\section{Background and Related Work}
\textbf{Large Language Models} such as language representation and generation models commonly use multiple layers of transformer encoders or decoders. Each transformer layer usually contains some form of multi-head attention (MHA) and fully-connected feed-forward networks (FFN). The MHA is made up of multiple self-attention heads \cite{Vaswani2017AttentionIA}, each of which has 3 sub-components: queries (\textbf{Q}), keys (\textbf{K}), and values (\textbf{V}). Equation \ref{eq1} shows the expression used to compute the \textit{attention} of each \textit{head}, where \textit{d} is the dimensionality of \textbf{K}. The output of the attention heads is concatenated and fed into the FFN. 
\begin{equation}
 Attention(Q,K,V) = \textit{softmax} \left(\frac{QK}{\sqrt{\textit{d}}}\right)V
  \label{eq1}
 \end{equation} \\
Attention, while simple, has proven to be incredibly robust as it allows models to scale to hundreds of layers, hundreds of attention heads \cite{Brown2020LanguageMA}, and seemingly most modalities \cite{Chen2021CrossViTCM} \cite{Arnab2021ViViTAV}. Despite its generalization ability, attention-based models are also brittle as removing less than 0.0001\% of parameters can cause complete model collapse \cite{Kovaleva2021BERTBO}.  \\
\textbf{Unstructured pruning} compresses a model by removing individual weights from a network by setting them to zero. Pruning methods commonly remove weights based on their saliency to the network and, to avoid model collapse, usually do so gradually while fine-tuning the remainder of the weights. Since it is difficult to quantify the true saliency of weight concerning a network, zeroth, first, and second-order estimation methods exist to approximate saliency.\\
Zero-order methods use weight magnitudes as a proxy,  \textit{i.e.}., remove the smallest weights without evaluating the impact of their removal on model accuracy. These approaches are prevalent for Convolution Neural Networks \cite{Han2015ADN} and have recently been successful for LLM \cite{Gordon2020CompressingBS} \cite{Chen2020TheLT} \cite{Zafrir2021PruneOF}. First-order methods like Movement Pruning \cite{Sanh2020MovementPA} use a gradient-based approximation to remove the weights moving toward zero. Second-order methods like OBS \cite{Singh2020WoodFisherES} estimate the impact of individual weight removal via approximations of second-order derivatives and use it as a proxy for saliency.\\
Using unstructured and semi-structured pruning has proved to be a convenient way of compressing LLM for efficient inference and decreased model size. For example, a BERT-Base-Uncased model which has had 90\% of its parameters pruned runs inference $\sim\!4.5$ times faster and is $2.75$ times smaller with no drop in accuracy \cite{Kurti2022TheOB}. If this compression leverages additional methods like quantization and layer dropping, inference speed can improve by 28.75 times, and model size can drop by $\sim\!19$ times. \\
While many successful compression approaches exist, Transformer models are fragile~\cite{DBLP:journals/corr/abs-2105-06990}, as minor perturbations can lead to model collapse. Moreover, there is a lack of understanding of the generality and robustness of compressed LLMs when transferring them to different tasks in an application domain, a gap our work attempts to fill. 
\section{Sparse*BERT: General Sparse Models Can Adapt to New Domains}
We can formulate the Sparse*BERT model as $\theta^*$, which can approximate the accuracy of the dense model $\theta$ and does not suffer from model collapse when transferred to new domains. The architecture of Sparse*BERT matches BERT, but a portion of its weights are pruned and masked to avoid future updates. To ensure that our sparse model can approximate the accuracy of the dense model, we leverage Knowledge Distillation between the outputs of the dense and sparse networks.   \\
Following the success of Zafrir et al. \cite{Zafrir2021PruneOF}, we leverage Gradual Magnitude Pruning (GMP) as shown in algorithm \ref{algo:gmp}. Pruning models using gradual magnitude pruning have been shown to be quite effective and easy to use. At its core, the goal of GMP is to remove weights slowly enough that the network can learn to recover after each pruning step.  \\
To ensure that we can maximize the inference speedups, we prune each set of components in the network independently. The structure in the model graph groups these components, so the individual feed-forward layers, the queries, or keys, but not individual self-attention heads.
\begin{algorithm}
\caption{Uniform Gradual Magnitude Pruning}
\label{algo:gmp}
\begin{algorithmic}
{\small
\Require $\theta_0$ a pretrained neural network, $\theta_t$ a dense pretrained neural network (distillation teacher), $D$ a training dataset, $N$ number of pruning steps, $\sigma$ weights to prune at each pruning step
\Ensure a pruned neural network
\For{$x$ in 1 to $N$ }
    \State $ \theta^*  \gets \theta_0$
    \For{component in  $\theta*$}
        \State $w\phantom{,} \gets sort(component)$
        \State $\theta^* \gets prune(\theta^*, w, \sigma)$ 
    \EndFor
    \State $\theta^* \gets train(\theta^*,\theta_t,D)$
    
\EndFor \\
return $\theta^*$
}
\end{algorithmic}
\end{algorithm}
\section{Experiments}
To assess how well-pruned models can transfer to new domains and determine the optimal stage of pruning (pretraining, domain transfer, or fine-tuning), we evaluate the accuracy of 10 Biomedical biomedical datasets. Our experiments fix the training parameters for all the transfer tasks and vary the stage used for pruning (no pruning, pretraining, domain transfer, fine-tuning) and domain-specific pretraining.
\subsection{Datasets}
In all of our experiments, we focus on English-only biomedical datasets/corpora. Each dataset we use is unmodified from its described use in the literature, and its associated and prescribed metrics are used.\\
\subsection{Finetuning Datasets}
\begin{table*}[!htb]
    \centering
    \scalebox{0.9}{ 
    \begin{tabular}{c|c|c|ccc|c}
    \toprule 
    Dataset & Domain & Task Type & Training Size & Testing Size & Validation Size & Evaluation Metric \\
    \midrule
    BC5-Chem & Medical & Entity Recognition & 5203 & 5347 & 5385 & F1 \\ 
    BC5-disease & Medical & Entity Recognition. & 4182 & 4244	 & 4424	 & F1 \\
    NCBI-disease & Medical & Entity Recognition & 5134 & 787 &	960 & F1 \\
    BC2GM	& Medical & Entity Recognition & 15197 &	3061 &	6325 & F1 \\ 
    JNLPBA	& Medical & Entity Recognition & 46750 &	4551 &	8662 & F1 \\	
    ChemProt & Medical &	Relation	& 18035 & 	11268	& 15745	& F1 \\
    DDI	& Medical & Relation	& 25296	&2496 &	5716 & F1 \\
    GAD & Medical&	Relation	&4261	& 535 &	534	 & F1 \\
    PubMedQA & Medical &	Question Answering &	450 &	50 &	500 &	Accuracy \\
    BioASQ	& Medical & Question Answering &	670	 & 75	& 140 &	Accuracy \\
    SQUAD & General & Question Answering & 87599& 10570 & N/A & F1  \\ 
    \bottomrule
    \end{tabular}
    }
    \caption{To understand how generalizable sparse models are, we evaluate a wide set of tasks that vary in difficulty, size, and desired output}
    \label{tab:dataset_descriptions}
    \vspace{-1.2em}
\end{table*}
\textbf{Pretraining Datasets.} To understand how the stage of pruning impacts model accuracy, we train models both pruned and dense models on the Medline/PubMed corpus and the combination of English Wikipedia \cite{wikidump} and The Book Corpus \cite{Zhu_2015_ICCV} datasets. \\
The combination of Wikipedia and Book Corpus datasets creates a common domain language dataset featuring ~3.3 billion words, which have become the backbone for general domain masked language modeling experimentation. \\
The MEDLINE/PubMed Corpus is a publicly available\footnote{https://www.nlm.nih.gov/databases/download/pubmed\_medline.html} text corpus made up of journal abstracts and documents of biomedical literature from around the world. The United States National Institute of Health updates the corpus daily and has been the primary resource used to train Biomedical LLMs like BioBERT \cite{Lee2020BioBERTAP} and PubMedBERT \cite{Gu2022DomainSpecificLM}. For our experiments, we extracted our corpus on January 2022, filtered and prepared the dataset for masked language modeling using the BioElectra's \cite{Kanakarajan2021BioELECTRAPretrainedBT} scripts\footnote{https://github.com/kamalkraj/BioNLP-Corpus}. This formatted PubMed corpus has 34,246,348 abstracts and 4.5 billion words \cite{Kanakarajan2021BioELECTRAPretrainedBT}. \\
\textbf{Finetuning Datasets.} We finetune pretrained models on 10 established Biomedical NLP datasets, encompassing 3 separate task types: Entity Recognition (ER), Relation Extraction (RE), and Question answering (QA). For ER, we use the BioCreative II Gene Mention Recognition (BC2GM), \cite{Smith2008OverviewOB}, BC5CDR Drug/Chemical (BC5-Chem), BC5CDR Disease (BC5-Disease) \cite{Li2016BioCreativeVC}, JNLPBA \cite{Collier2004IntroductionTT}, and NCBI Disease \cite{Dogan2014NCBIDC} datasets. For RE we use ChemProt \cite{Taboureau2011ChemProtAD}, Drug-Disease Interaction (DDI) \cite{HerreroZazo2013TheDC}, and Gene-Disease Associations (GAD) \cite{Becker2004TheGA} datasets. We leverage BioASQ task 7B \cite{Baker2016AutomaticSC} and PubMedQA \cite{Jin2019PubMedQAAD} for QA. In addition, we analyze the impact of the finetuning dataset's size on the optimal pruning stage using the non-biomedical QA SQUAD \cite{Rajpurkar2016SQuAD1Q} dataset. Details on the dataset size, evaluation metric, and domain can be found in Table \ref{tab:dataset_descriptions}.\\
\subsection{Models and Experimental Setup}
Our experiments focus on the popular BERT-base-uncased language model \cite{Devlin2019BERTPO}, an LLM composed of 12 transformer encoder layers with 110M parameters. We compare the performance of our sparse language models to the dense, task-tuned BioBERT model, which is also a 12 transformer layer model which features cased (BioBERT-Base-Cased) and uncased (BioBERT-Base-Uncased)variants. Following previous work, we do not prune the embedding layers of the network or any task-specific heads and focus on the $\sim\!85$ million parameters found in the encoder layers. To ensure that our experiments are reproducible \footnote{we will be releasing our code and models to allow reproducibility and extensibility} we use the popular open-source libraries SparseML \footnote{https://github.com/neuralmagic/sparseml} and Transformers \footnote{https://github.com/huggingface/transformers}. \\
\textbf{Model Pretraining}
Pretraining is when the model is trained on an unsupervised NLP dataset using a masked language modeling (MLM) approach \cite{Devlin2019BERTPO}. Pretrained models are fine-tuned on labeled task-specific datasets to optimize for task-specific accuracy.\\
Our experiments use existing dense pretrained models for BERT-base-uncased \cite{Devlin2019BERTPO} and PubMedBERT \cite{Gu2022DomainSpecificLM} and prune them using gradual magnitude pruning based on the corresponding dataset and MLM approach. \\
During model pretraining, we train for three epochs on 4 A100 GPUS using a batch size of 256 and a sequence length of 512, and, following early experiments and findings from Kurtic et al.\ and Zafir et al., we cycle the learning rate during pretraining and found cycling twice per epoch from 5e-4 to 0 to be most effective. First, we take the existing dense models and run this training setup for 3 epochs to ensure model convergence; then, taking these converged models, we retrain and apply gradual magnitude pruning over the first two epochs. During pruning, we start from an initial sparsity of $30\%$ and gradually prune to a final sparsity of 90\%, pruning 100 times an epoch. After model pruning, we continue to train for one additional epoch to ensure that the sparse model is converged. 
Based on early experiments, we find knowledge distillation beneficial. For all of our experiments in pretraining, we leverage well-trained dense teachers using a hardness of  0.5 and a temperature of 2. When pruning weights, their values are fixed to 0 and are masked to avoid future updates. This means that our experiments effectively evaluate the discovery of the most optimal sub-architecture.
\textbf{Model Finetuning}
Finetuning is the stage in which the model is trained on a supervised NLP dataset using a task-specific training regime. In this stage, one or many classification heads have connected the model, and these classification heads and the pretrained model are trained for optimal performance on an individual task.\\
We fix the training procedure across fine-tuning tasks to isolate the effects of task-specific hyperparameter tuning and pruning stages. Specifically, we train each model for ten epochs on a V100 GPU using a batch size of 16, a learning rate that linearly decays from 5e-5 and replicates using five random seeds for larger tasks and ten random seeds for smaller tasks. \\
\begin{table*}[htb!]
    \centering
    \scalebox{0.8}{ 
    \begin{tabular}{c|c|ccc|c}
    \toprule 
    Model & Pruning Stage & EE & RE & QA & Overall \\
    \midrule
    BERT-Base-Uncased & \makecell{None \\ Fine-tuning \\ Pretraining} & \makecell{\textbf{84.44}  \\ 76.12 \\ 80.84}  & \makecell{\textbf{86.86} \\ 85.36 \\ 85.54} & \makecell{61.97 \\ 65.39 \\ \textbf{68.44}} & \makecell{77.76 \\ 75.28 \\ \textbf{78.27}}\\
    \midrule
    BioBERT-Base-Uncased &  \makecell{None \\ Fine-tuning \\ Pretraining (Medical)\\ Pretraining (General)} & \makecell{85.96 \\ 63.53 \\ 82.50 \\ \textbf{86.36}} & \makecell{87.56 \\ 75.14 \\ 86.60 \\ \textbf{88.57}}  & \makecell{\textbf{68.33}\\  54.00 \\ 65.71 \\ 66.33} & \makecell{\textbf{80.62} \\ 66.34 \\ 78.27 \\ 80.42}  \\
    \bottomrule
    \end{tabular}
    }
    \caption{Overall results on the impact of task and dataset of model pruning. Models trained for the general domain and pruned on the general domain can transfer at equal or better accuracy. Question Answering is the notable outlier as its small dataset size benefits from the sparse models as their pruned architecture prevents overfitting on small datasets.}
    \label{tab:full_datasets}
    \vspace{-1.2em}
\end{table*}
We use the same setup for fine-tuning already pruned models and when applying gradual magnitude pruning during fine-tuning (pruning on the fine-tuning stage). For pruned models, we preserve the sparsity patterns. When pruning models during fine-tuning, we fine-tune the dense model for two epochs, prune over the preceding six epochs, and stabilize the pruned network for two epochs. In our early experiments and matching prior findings \cite{Zafrir2021PruneOF}, we find that when pruning models on transfer tasks, accuracy is best when the learning rate cycles. Cycling only occurs when pruning during fine-tuning, and the learning rate cycles at epochs 2 (start of pruning) and 8 (end of pruning). Unlike previous work, we do not find a significant effect in accuracy improvement by leveraging knowledge distillation on the fine-tuning task. As a result, we do not use knowledge distillation during fine-tuning.
\subsection{Experimental Results}
\begin{table*}[htb!]
    \centering
    \scalebox{0.6}{ 
        \begin{tabular}{cc|ccccc|ccc|cc}
    \toprule 
    Model & Pruning Stage & BC5-Disease & BC5-chem & NCBI-disease & BC2GM & JNLPBA & ChemProt & DDI & GAD & \makecell{PubMedQA \\ Accuracy} & \makecell{BioASQ \\ Accuracy} \\
    \midrule
    Training dataset size & N/A& 4182 & 5203 & 5134 & 15197 & 46750 & 18035 & 25296 & 4261 &  450 & 670 \\
    \midrule
    BERT-Base-Uncased & \makecell{None \\ Fine tuning \\ Pretraining}  & \makecell{80.60 \\ 69.87 \\ 75.35} & \makecell{91.23 \\ 81.72 \\ 87.83} & \makecell{\textbf{85.66}\\ 75.57 \\ 81.75 } & 
    \makecell{81.97 \\ 74.27 \\ 77.12} & \makecell{81.56 \\ 79.57 \\ 81.24}  &  \makecell{88.19 \\ 85.41 \\ 86.13} & \makecell{94.35 \\ 92.72\\ 92.73} & \makecell{78.05 \\ 74.88 \\ 77.77}  & \makecell{47.46 \\52.67 \\ 50.00} & \makecell{76.65 \\ 78.11 \\ 82.89} \\
    
    \midrule
    BioBERT-Base-Uncased &  \makecell{None \\ Fine tuning \\ Pretraining(General) \\ Pretraining(Medical)}  &\makecell{83.195 \\ 66.34 \\\textbf{83.32} \\ 80.56 } &  \makecell{93.63 \\ 66.60 \\ \textbf{93.81} \\92.17 } & \makecell{83.46 \\ 52.69 \\ 84.15 \\ 78.84}&  \makecell{86.05 \\ 59.71 \\\textbf{87.04} \\ 81.11} &  \makecell{\textbf{84.10} \\62.15 \\ 81.84 \\ 81.81 } &   \makecell{90.66 \\ 80.28 \\ \textbf{90.71} \\ 88.02} &  \makecell{95.01 \\ 87.60  \\ \textbf{95.02} \\ 93.99} &  \makecell{77.02 \\ 57.55 \\ \textbf{79.90} \\ 77.77} &   \makecell{\textbf{54.00}\\ \textbf{54.00} \\ 51.50 \\ 49.14} &  \makecell{\textbf{82.67}\\ \textbf{82.67} \\ 81.17 \\ 82.28} \\
    \bottomrule
    \end{tabular}
    }
    \caption{Performance on Complete set of tasks. Except for question-answering tasks and NCBI-Disease, the SparseBioBERT outperforms all other models, including BioBERT, indicating that sparse architectures can be transferred to new domains and use cases without additional optimization. }
    \label{tab:full}
    \vspace{-1.2em}
\end{table*}
When we evaluate results on an average of task-specific scores as shown in Table \ref{tab:full_datasets}, we can see that the SparseBioBERT model performs on par with its unpruned counter-part and outperforms it on relation extraction and entity extraction. Results on individual tasks can be found in table \ref{tab:full} and further consistently show how SparseBioBERT approximated BioBERT. When pruned models do not transfer to the biomedical domain, they can perform much worse than the unpruned models, as shown by the sizable gap between the pruned and dense BERT-base-uncased models. This result, coupled with the performance of SparseBioBERT, makes us believe that pruned models can adapt to new domains like unpruned models but require additional domain-specific pretraining to ensure performance. \\
We believe these results prove that models pruned during general domain data can remove large portions of the model without affecting the ability to transfer to new domains or tasks. Unlike pruning on specific data domains and tasks, general domain data pruning can preserve accuracy invariant to task and domain. \\
Unexpectedly, when evaluating biomedical QA, we improve accuracy by pruning, but only on a general domain language model pruned downstream or the general domain Sparse*BERT. \\ We attribute this to the regularizing effect that pruning can have, and it likely helps in overfitting small datasets, PubMedQA and BioASQ. Tasks. Finding those models pruned outperform all others, and we believe the regularization provided by pruning can prevent overfitting on these small datasets.  \\
Our results also indicate that it is optimal to prune with general domain data and transfer it to new tasks and domains for optimal performance. Regardless of their domain expertise, BERT and BioBERT both see huge losses in accuracy when pruned on the downstream tasks, and these same losses are not found in the model pruned during pretraining. Surprisingly, the model pruned on the general domain data outperforms when pruned on biomedical domain-specific language modeling. This gap is nearly 4 points on entity extraction and 2 points overall, almost more significant than the gap between the BERT and BioBERT. \\
When we evaluate the impact of pruning on individual tasks pruning in fine-tuning stage) as shown in Table \ref{tab:full}, we can see that pruning is quite sensitive to the dataset task. Looking at the large datasets like JNLPBA in Table \ref{tab:full}, there is nearly no distinction in pruning during pretraining or fine-tuning. On the other hand, small datasets like NCBI and GAD see a large accuracy loss from models pruned during fine-tuning.  \\
\section{Impact of Training Data Size}
\begin{figure}
   \centering
    \includegraphics[scale=0.5
    ]{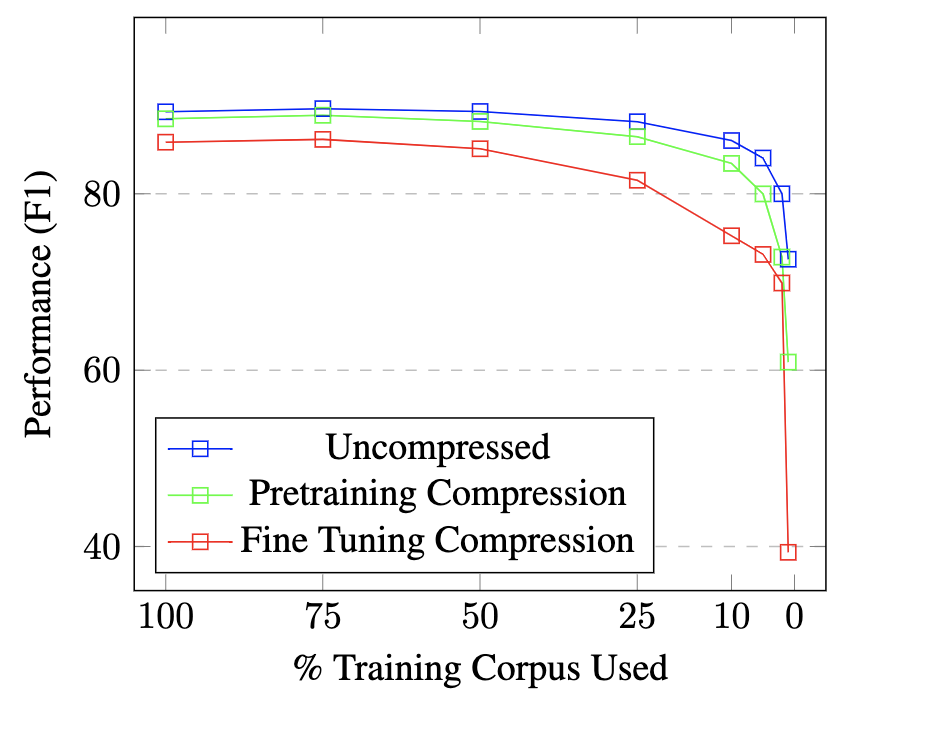}
    \vspace{-1.2em}

    
%
\caption{Impact of varying the training data size with pruned and dense models showcases how pretraining pruning has similar accuracy to the dense models.}
\label{fig:squad_data_size}

\end{figure}

Noting that there is a significant variation in dataset size in the biomedical NLP tasks, we leveraged a dataset well studied in pruning, SQUAD \cite{Rajpurkar2016SQuAD1Q}, and performed variations to the training data size. Starting with the initial training dataset size, 88,000 items, we decreased the size to $75\%$,$50\%$,$25\%$,$10\%$,$5\%$,$2\%$,$1\%$ and evaluated the impact to performance. We compared the dense BERT, Sparse*BERT, and pruning BERT during fine-tuning. The sparse models each have $90\%$ unstructured sparsity on the encoder portion of the LLM. \\ 
Each experiment was performed with five random seeds, using a batch size of 12, and trained for 30 epochs with a learning rate decaying linearly from 5e-5 to 0. Each model's training uses knowledge distillation from a dense teacher with the hardness set to 0.8 and the temperature to 2.0. For the model pruned during fine-tuning, we cycle the learning rate at the beginning of pruning(2nd epoch) and the end of pruning(20th epoch). We evaluate model performance on the F1 score on the unaltered dev portion of the SQUAD dataset to avoid losses in evaluation sensitivity.\\
Figure \ref{fig:squad_data_size} and table \ref{tab:squad_dataset} demonstrate that models pruned during finetuning are not robust to variations in data size. Model performance decays slowly from 85 to 80 until the training data is decreased by $75\%$, but it quickly becomes nearly unusable when it becomes smaller than that. The same cannot be said about the dense or the Sparse*BERT model, as they see virtually identical losses in quality from 
\begin{table*}[htb!]
    \centering
    \scalebox{0.9}{ 
    \begin{tabular}{|c|ccc|}
    \toprule 
     Training Data Portion& BERT-BASE-Uncased  &  Sparse*BERT & Finetune Pruning \\
    \midrule
    100 & 89.30 & 88.50 & 85.852 \\ 
    75 & 89.65 & 88.91 & 86.17 \\
    50 & 89.33 & 88.21 &  85.11 \\
    25 & 88.17 & 86.48 & 81.5 \\
    10 & 86.04 & 83.45 & 75.24 \\
    5 & 84.06 & 79.99 & 73.13 \\
    2 & 80.00 & 72.83 & 69.88 \\
    1 & 72.58 & 60.91 & 39.35 \\
    \bottomrule
    \end{tabular}
    }
    \caption{Model accuracy as measured by F1 on dev portion of SQUAD compared to model type and the impact of training data size. Sparse models are not as sample efficient as their dense counterparts, but Sparse*BERT performance matches the dense model much more than the model pruned downstream.}
    \label{tab:squad_dataset}
    \vspace{-1.2em}
\end{table*}
\section{Limitations}
Our approach is limited in the computational time required to generate a general sparse LLM and the diversity of types of LLMs that we explore. \\
Regarding computational expense, training a sparse model requires negligible additional resources, which is tractable for models with a hundred million parameters and a few billion tokens, not for billion parameter models commonly discussed. \\
Our current explorations have been limited to monolingual LLMs trained in English. It is unclear how well sparse architectures will perform in a multi-lingual setting, and we expect degradation in language quality to be anything but equal across all languages. \\
\section{Conclusion}
In this work, we have introduced Sparse*BERT, a pruned LLM which builds on successful pruning algorithms and demonstrates its ability to transfer to new domains and tasks without additional hyperparameter search. Our experiment demonstrates how well Sparse*BERT can transfer to the biomedical domain to become SparseBioBERT. SparseBioBert can match the performance of BioBERT with $\frac{1}{10}$ of the parameters and no additional task-specific optimization. 
In future work, we seek to expand Sparse*BERT to new legal, financial, and medical domains. Furthermore, we like to continue our work on more complex models studying how sparsity impacts multilingual and multi-task models. In particular, we seek to understand how structured and unstructured approaches in compression relate to the curse of multilingualism.
\bibliography{anthology,custom}
\bibliographystyle{acl_natbib}
\end{document}